\documentclass[11pt]{article}

\usepackage[T1]{fontenc}
\usepackage[utf8]{inputenc}
\usepackage[margin=1in]{geometry}
\usepackage{mathpazo}
\usepackage[scaled=0.92]{helvet}
\usepackage{microtype}
\usepackage{amsmath}
\usepackage{amssymb}
\usepackage{graphicx}
\usepackage{booktabs}
\usepackage{multirow}
\usepackage{array}
\usepackage{tabularx}
\usepackage{url}
\usepackage{caption}
\usepackage[numbers,sort&compress]{natbib}
\usepackage[colorlinks=true,linkcolor=blue,citecolor=blue,urlcolor=blue]{hyperref}

\usepackage{epstopdf}
\title{A Lightweight Multi-View Approach to Short-Term Load Forecasting}

\author{%
  Julien Guité-Vinet\thanks{Corresponding author: \texttt{guite-vinet.julien@courrier.uqam.ca}.} \quad
  Alexandre Blondin Massé \quad
  Éric Beaudry \\[4pt]
  \small Université du Québec à Montréal, Département d'informatique\\
  \small 201 Av. du Président-Kennedy, Montréal, QC H2X 2J5, Canada
}
\date{}

\begin{document}

\maketitle

\begin{abstract}
Time series forecasting is critical across domains such as energy, finance, and meteorology, where accurate predictions enable informed decision-making. While transformer-based and large-parameter models have recently achieved strong results, their complexity can increase the parameter count and make comparisons under limited data more difficult. In this paper, we address short-term load forecasting with a multi-view approach combining single-value embeddings and a scaled time-range input to capture temporally relevant features efficiently. The proposed lag set includes recent observations together with daily, weekly, and yearly lags; the weekly and yearly lags remain anchored on observed values during forecasts of up to one week. We introduce an embedding dropout mechanism that provides a qualitative analysis of the effect of individual features on the forecast. Evaluated under a strictly chronological protocol, the proposed model uses 33,605 trainable parameters, compared with 104,408--346,818 for the reported baseline checkpoints. Its forecasting accuracy varies across datasets and horizons: it is not the best-performing model in every setting, but provides a compact alternative to the larger neural baselines. The approach is evaluated on three public datasets, including scenarios with corrupted sensor readings.

\end{abstract}

\section{Introduction}
Time series forecasting is a fundamental problem with broad applications, including energy demand management, financial market analysis, and meteorological predictions.
Reliable forecasts are essential for resource allocation, operational planning and informed decision-making.
Traditional statistical models offer interpretability and robustness~\cite{xgboost,clements,hyndman}, but often require extensive feature engineering and may struggle to capture complex temporal dependencies.
In contrast, deep learning models—particularly recurrent neural networks~\cite{lstm-prevdem,deepar} and transformer-based architectures~\cite{informer}—can learn rich representations from raw sequences and model complex temporal dependencies.

Despite these advances, time series data present unique challenges.
Long historical sequences may contain outdated information, and large-parameter models can overfit when data are limited or exhibit non-stationary patterns.
Consequently, reducing the parameter count while retaining useful predictive accuracy is an important design objective.

In this work, we address short-term load forecasting with a compact multi-view architecture.
The proposed method encodes each view into a compact embedding, combining single-value embeddings with standard embedding strategies.
A scaled time-range input restricts the model to temporally relevant observations, so that the seasonal and the short-term patterns are captured within a small parameter footprint.
An embedding dropout mechanism prevents the model from relying on any single view, which improves robustness and supports interpretation.

The contributions are threefold:
\begin{enumerate}
	\item \textbf{Efficient multi-view encoding.} Single-value embeddings encode each view with one scalar weight, substantially reducing the number of trainable parameters relative to the other embedding variants. The resulting accuracy depends on the dataset and embedding strategy.
	\item \textbf{Embedding dropout for robustness.} A dropout applied at the granularity of a view limits the dependence on any single covariate, and serves as an ablation mechanism that exposes the contribution of each view to the forecast.
	\item \textbf{A structured lag set.} The scaled time-range input combines recent observations with daily, weekly, and yearly lags. Its weekly and yearly entries remain anchored on observed values during forecasts of up to one week, a structural property that we evaluate across three public load datasets.
\end{enumerate}

The proposed model contains 33,605 trainable parameters in the reported experiments, compared with 104,408--346,818 for the baseline checkpoints. The remainder of this paper is structured as follows.
Section~\ref{sec:formulation} formally defines the forecasting problem.
Section~\ref{sec:sota} reviews related work.
Section~\ref{sec:preliminaries} introduces background on transformers and embeddings.
Section~\ref{sec:proposed-approach} details the proposed approach.
Section~\ref{sec:experimental-settings} presents the experimental setup, followed by results in Section~\ref{sec:experimental-results}.
Finally, Sections~\ref{sec:discussion} and~\ref{sec:conclusion} provide discussion and conclusions.

\section{Problem Formulation}\label{sec:formulation}
We formulate the task as a regression problem in which future values of a time series are predicted from its history and associated exogenous features.
More specifically, let $\mathcal{Y}$ denote the target series, $\tau$ the forecasting horizon, and $m$ the length of the context window ending at time $t$.
The objective is to minimize a loss that measures the forecasting performance of model $\mathcal{M}$.
At time $t+1$, the model produces $\tau$ forecasts from the historical target $\mathcal{Y}[t-m:t]$ and the exogenous series $\{\mathcal{X}[t-m:t]_f\}_{f=2}^{F}$.
We call both the target series and each exogenous series a view, giving a total of $F$ views.
Here, a view denotes an exogenous variable together with its separately learned projection, rather than a distinct observation modality in the multi-view learning literature; this definition allows each view to be removed independently at inference.
Formally, this optimization problem is expressed as:
\begin{equation}
	\begin{split}
		\hat{\mathcal{Y}} &= \mathcal{M}_{\theta}\left( \mathcal{Y}[t-m:t], \{\mathcal{X}[t-m:t]_f\}_{f=2}^{F} \right), \\
		\theta^{*} &= \operatorname*{arg\,min}_{\theta} \; \mathcal{J}\left( \mathcal{Y}[t+1:t+\tau], \hat{\mathcal{Y}} \right),
	\end{split}\label{eq:task}
\end{equation}
where $\theta$ denotes the parameters of $\mathcal{M}$, $\mathcal{J}$ the loss function, $\hat{\mathcal{Y}}$ the $\tau$ forecasts, and $\mathcal{Y}[t+1:t+\tau]$ the observed future values against which they are evaluated.

Many loss functions can be used to evaluate the performance of the model, including the Mean Squared Error, Mean Absolute Error, and Mean Absolute Percentage Error (MAPE).
The MAPE loss is defined as follows:
\begin{equation}\label{eq:mape}
	\text{MAPE} = \frac{100}{n} \sum_{i=1}^{n} \left| \frac{y_i - \hat{y}_i}{y_i} \right|
\end{equation}
where $n$ is the number of data points, $y_i$ is the true value and $\hat{y}_i$ is the forecasted value.

MAPE is the only metric reported in this study and is used as the training loss.
Because it is scale free, it can be compared across systems of different installed capacities.
It is also the metric most commonly reported for short-term load forecasting and is interpretable as a relative deviation.
Its known instability when the target approaches zero does not arise for an aggregated load.
It remains asymmetric, however, penalising overestimation more heavily than underestimation.
Table~\ref{tab:notation} summarises the notation used throughout the paper.

\begin{table}[t]
  \centering
  \small
  \caption{Notation used throughout the paper.}
  \label{tab:notation}
  \begin{tabularx}{\textwidth}{@{}l>{\raggedright\arraybackslash}X l>{\raggedright\arraybackslash}X@{}}
    \toprule
    \multicolumn{2}{l}{\textbf{Problem formulation}} & \multicolumn{2}{l}{\textbf{Embedding}} \\
    $\mathcal{Y}$ & target load series & $d$ & embedding dimension of a single value \\
    $\mathcal{X}$ & matrix stacking the $F$ views & $d_{\mathrm{model}}$ & width of the aggregated representation \\
    $\mathcal{M}_{\theta}$ & forecasting model of parameters $\theta$ & $W_f$, $b_f$ & projection and bias of the view $f$ \\
    $\mathcal{J}$ & loss function & $w_f$ & scalar weight of the view $f$ under SVE \\
    $\tau$ & forecasting horizon & $z_{l,f}$ & embedding of the entry at lag $l$ for the view $f$ \\
    $m$ & context window length & $z_f$ & embedding of the whole view $f$ \\
    $F$ & number of views, target included & $Z$ & collection of the view embeddings \\
    $t$ & anchor time of a forecast & $p$ & embedding dropout rate \\
    $n$ & number of data points in the MAPE & $r_f$ & Bernoulli mask of the view $f$ \\
    \addlinespace
    \multicolumn{2}{l}{\textbf{Scaled time-range input}} & $\hat{z}_f$ & embedding of the view $f$ after dropout \\
    $\mathcal{T}$ & set of datetimes of the dataset & $\tilde{Z}$ & aggregated representation \\
    $\mathcal{T}_{\mathrm{valid}}$ & datetimes admitting every lag & \multicolumn{2}{l}{} \\
    $T$ & number of rows available in $\mathcal{X}$ & \multicolumn{2}{l}{\textbf{Encoder--decoder}} \\
    $\mathcal{L}$ & lag set & $E$ & transformer encoder \\
    $l$ & a lag of $\mathcal{L}$ & $H$ & encoded states \\
    $f$ & a view, $1 \le f \le F$ & $D^{\ell}$ & decoder layer $\ell$ \\
    $x_{t-l}$ & row of $\mathcal{X}$ observed at time $t-l$ & $N$ & number of decoder layers \\
    $\tilde{X}_t$ & scaled time-range input at time $t$ & $U^{\ell}$ & values fed to the decoder layer $\ell$ \\
    $k$ & forecast step, $k \in \{1, \dots, \tau\}$ & $Q^{\ell}$, $K^{\ell}$, $V^{\ell}$ & query, key and value matrices \\
    \addlinespace & & $W_Q^{\ell}$, $W_K^{\ell}$, $W_V^{\ell}$ & attention projection matrices \\
    \multicolumn{2}{l}{\textbf{Attention background}} & $\tilde{Y}^{\ell}$ & output of the decoder layer $\ell$ \\
    $d_k$ & dimension of the key vectors & $\hat{Y}$ & forecast vector, one entry per lag \\
    $S$ & length of the input sequence & \multicolumn{2}{l}{} \\
    \bottomrule
  \end{tabularx}
\end{table}

\section{Related Work}\label{sec:sota}
Forecasting accuracy has been the focus of extensive research, leading to the development of various methods, including autoregressive models, moving average models, and exponential smoothing techniques~\cite{hyndman}.
Statistical approaches, such as SARIMAX, have consistently demonstrated reliability across multiple application domains, and remain in operational use for short-term electrical load forecasting~\cite{arimax}, where
models such as XGBoost have emerged as powerful alternatives, achieving competitive performance in both public forecasting competitions and real-world deployments~\cite{forecast-competition}.

In parallel, machine learning-based techniques have gained significant traction.
Deep learning has further advanced the field of load forecasting, with architectures such as recurrent neural networks, long short-term memory networks, convolutional neural networks and transformer-based networks being widely explored~\cite{tide,convolutional-prevdem,lstm-prevdem,nbeats,deepar,informer}.
These models excel in capturing complex temporal dependencies, thereby reducing the need for extensive feature engineering traditionally required by classical methods.
Section~\ref{sec:preliminaries} reviews most of these architectures in more detail.

However, opinions among researchers and practitioners differ regarding these claims.
Some have questioned the assumption of improved accuracy, presenting evidence that simpler models outperform more complex approaches with less cost and complexity~\cite{forecast-competition,m3-competition}.
On long-horizon benchmarks, a linear model applied to a decomposed series matches or exceeds several transformer architectures~\cite{dlinear}.

We argue that this may be due to the fact that time series datasets often lack the scale needed for deep learning methods to perform well and
that reported gains can be difficult to compare across studies.
While benchmark collections have become available, evaluation protocols still vary, and comparisons between studies adopting different splitting strategies should be interpreted with care.
Shuffling the observations before splitting, or drawing folds at random, places future values in the training set and inflates the reported accuracy; this form of leakage is documented across quantitative fields and is a recurring pitfall of forecast evaluation~\cite{forecast-pitfalls,ml-leakage}.
For this reason, we report all results under a single chronological protocol, described in Section~\ref{subsec:training-testing}.

Subsequent architectures answer that critique by changing what attention is applied to.
PatchTST attends to patches of the series rather than to individual timesteps~\cite{patchtst}, iTransformer attends across variates instead of timesteps~\cite{itransformer}, and TimesNet folds the series into two dimensions to model intra-period and inter-period variation~\cite{timesnet}.
All of them take a contiguous and uniformly sampled look-back window as input, an assumption from which the present work departs.

Short-term load forecasting has long proceeded otherwise.
As stated in~\cite{clements}, a key ingredient of good forecasting is the effective use of lagged information by allowing for interaction between seasonal patterns and intra-day dependencies.
Which lags carry that information is already established for load demand: the series exhibits a daily and a weekly cycle~\cite{double-seasonal}, to which an annual cycle is added when the sample spans several years~\cite{load-forecasting-review,triple-seasonality}.
Selecting a few lags rather than a full window is likewise established practice, semi-parametric models of the load relying on a small set of lagged loads and temperatures~\cite{semiparametric-load}.

\section{Preliminaries}\label{sec:preliminaries}
In this section, some preliminary concepts are introduced to understand the proposed approach.
Section~\ref{subsec:transformer} recalls the attention mechanism on which the proposed encoder is built, and Section~\ref{subsec:encoder-decoder} reviews the encoder--decoder architectures used as baselines.
Surveys cover deep forecasting models~\cite{timeseries-forecasting-survey} and transformer variants for time series~\cite{transformer-survey}.

\subsection{Transformers}\label{subsec:transformer}

A transformer is a deep learning model that has been widely used, first in natural language processing and more recently in time series forecasting.
It is based on an attention mechanism that allows the model to capture long-range dependencies in the data~\cite{attention}.
It allows the model to focus on different parts of the input with the help of an attention mechanism measuring the importance of each data point in the input sequence.

To achieve this, the model utilizes three matrices: the query matrix $Q$, the key matrix $K$, and the value matrix $V$,
with $Q$ representing the input for which relevance is assessed, $K$ containing the features used for comparison, and $V$ holding the data to be aggregated based on the attention scores.
Formally, the attention mechanism is defined as:
\begin{equation}
	\text{Attention}(Q, K, V) = \text{softmax}\left(\frac{QK^\top}{\sqrt{d_k}}\right)V\label{eq:equation2}
\end{equation}
where $d_k$ is the dimension of the key vectors, whose square root rescales the dot products to keep the softmax in a well-conditioned range.

The standard attention mechanism exhibits a time and memory complexity of $\mathcal{O}(S^2)$ per layer, where $S$ is the length of the input sequence.
Several studies have proposed methods to mitigate this issue by improving the computational efficiency of self-attention.
Notably, Informer~\cite{informer}, Reformer~\cite{reformer}, and LogSparse Transformer~\cite{logsparse} achieve a reduction in complexity to $\mathcal{O}(S \log S)$, albeit with a slight trade-off in accuracy.

\subsection{Encoder-Decoder models}\label{subsec:encoder-decoder}
Many popular models are designed to encode input representations into hidden states $H$ and decode output representations from $H^t=\{h^t_1, \dots, h^t_m\}$.
The inference follows a dynamic decoding process, where the decoder updates the hidden state $h^t_{k+1}$ based on the previous state $h^t_k $ to forecast $y^t_{k+1}$~\cite{tide,nbeats,deepar,informer}.

Time-series Dense Encoder (TiDE)~\cite{tide} encodes the past of a time series along with dynamic covariates and static attributes using dense multi-layer perceptron to achieve accurate forecasting.
N-BEATS~\cite{nbeats} employs three types of stacks, with one stack dedicated to modeling the trend, another to capturing periodicity, and a third to handling the residuals, all interconnected through residual connections to enhance the representation of temporal dependencies.
DeepAR~\cite{deepar} is an autoregressive recurrent neural network with probabilistic outputs, effectively capturing temporal dependencies for univariate time series forecasting.
Informer~\cite{informer} uses a probability-based attention mechanism to focus computation on the most informative queries and progressively reduces the sequence length between encoder layers.
Its generative decoder then produces the complete forecast in a single forward pass.
The Temporal Fusion Transformer (TFT)~\cite{tft} combines recurrent processing with attention and variable-selection mechanisms to encode static, past and future covariates, then uses a gated decoder to produce multi-horizon forecasts.
PatchTST~\cite{patchtst} encodes the input into patches with a Transformer encoder and produces forecasts through a prediction head, thus implementing decoding implicitly rather than with a separate decoder.
DLinear~\cite{dlinear} decomposes the input into trend and seasonal components and maps them directly to the forecast through linear projections, without an explicit encoder--decoder structure.

\subsection{Embeddings}\label{subsec:embeddings}
Embeddings encode data into a structured representation within a continuous vector space, where semantically similar features are positioned closer together while dissimilar features remain farther apart.
The objective is to generate embeddings that effectively preserve these relationships, ensuring that similar entities are mapped to proximate regions within the embedding space.
High-quality embeddings are essential for achieving accurate forecasts and robust model performance~\cite{multi-embeddings-witschard}.
Common strategies for combining embeddings include principal component analysis kernels~\cite{pca-kernel}, graph-based embedding methods~\cite{graph-embedding}, or transforming input features into a higher-dimensional space followed by aggregation via summation or concatenation~\cite{camul}.
The approach adopted here belongs to the last family, and departs from it by the width allotted to each view: a view is encoded by a single scalar rather than by a vector, which keeps the representation compact.

\section{Proposed Approach}\label{sec:proposed-approach}
In this section, we describe our approach to short-term load forecasting using a transformer-based model that embeds each view, as defined in Section~\ref{sec:formulation}, into a continuous vector.

The lag set proposed here follows the seasonal structure recalled in Section~\ref{sec:sota}, retaining one representative of each cycle together with the most recent hours, rather than the contiguous window that a fixed context length would impose.
The short lags serve a distinct purpose: sampled hourly from $6h$ to the forecast origin, they describe the level and the slope of the series in the immediate past, which the seasonal lags leave unresolved because they are spaced too far apart.
Between these two groups, the spacing doubles from $6h$ to $12h$ and from $12h$ to $1d$, so that the resolution of the input decreases with the distance to the forecast origin.
This progression is a design choice rather than a tuned one, and the sensitivity of the model to the number of short lags is left to further study.

Hence, instead of using an equally spaced time input, we propose to use a scaled time-range input,
\emph{i.e} from a starting datetime forecast, we build the input by using the time series lagged information of the previous year (y), previous week (w), previous day (d), previous half-day (12h), the last six hours and the current observation (0h).
This reduces the number of input positions while retaining the attention mechanism over the selected views and lags.

Let $\mathcal{X} = \left[\, \mathcal{Y}, \mathcal{X}_2, \dots, \mathcal{X}_F \,\right] \in \mathbb{R}^{T \times F}$ be the matrix stacking the $F$ views over $T$ hourly timesteps, the target series being the first one,
a list of lags be $\mathcal{L}$,
the set of observed datetimes be $\mathcal{T}$,
and
a valid datetimes set be $\mathcal{T}_{\mathrm{valid}}$ as a subset of $\mathcal{T}$ where
\begin{equation}
	\begin{aligned}
		\mathcal{L}=\{1y, 1w, 1d, 12h, 6h, 5h, 4h, 3h, 2h, 1h, 0h \}, \\
		\mathcal{T}_{\mathrm{valid}} = \{\, t \in \mathcal{T} : t - \max(\mathcal{L}) \ge \min(\mathcal{T}) \,\},
	\end{aligned}
\end{equation}
with $\min(\mathcal{T})$  and  $\max(\mathcal{L})$ denoting respectively the earliest datetime in $\mathcal{T}$ and the largest lag duration in $\mathcal{L}$.
The null lag $0h$ denotes the observation at the forecast origin, which is available at the time the forecast is issued.

The resulting scaled time-range input is the stacked matrix
\begin{equation}
	\tilde{X}_t =
	\begin{bmatrix}
		x_{t-l}
	\end{bmatrix}_{l \in \mathcal{L}}
	\in \mathbb{R}^{|\mathcal{L}| \times F}, \quad \forall t \in \mathcal{T}_{\mathrm{valid}},
\end{equation}
where $x_{t-l} \in \mathbb{R}^{F}$ is the row of $\mathcal{X}$ observed at time $t-l$, so that $\tilde{X}_t$ retains $|\mathcal{L}|$ rows out of the $T$ available ones.

An important consequence of this construction concerns multi-step forecasting.
The model decodes one step at a time, so that predicting the step $t+k$ for $k \in \{1, \dots, \tau\}$ relies on the entry $x_{t+k-1-l}$ for each lag $l \in \mathcal{L}$.
That entry is an observed value, rather than a previously generated forecast, as long as the anchoring condition $k \le l + 1$ holds.
Every lag $l \ge \tau - 1$ therefore remains anchored on ground truth over the entire horizon, and only the shorter lags are progressively replaced by the predictions of the model.
For a seven-day horizon, the weekly and yearly lags stay anchored at every step; for a one-day horizon, the daily lag is anchored as well.

This property is the primary motivation behind the proposed lag set.
It supplies the decoder with at least two undegraded references at any horizon up to one week, which can help limit error accumulation during autoregressive continuation without requiring a contiguous context window of the same span.
A model fed with a contiguous window of the same span cannot offer the same property, because its oldest entry is the one that leaves the observed region first, and the yearly seasonality lies entirely outside its receptive field.

Each entry $\tilde{x}_{l,f} \in \mathbb{R}$ of $\tilde{X}_t$, taken at lag $l \in \mathcal{L}$ for the feature $f$, is projected to a $d$-dimensional embedding
\begin{equation}
	\begin{aligned}
		z_{l,f} & = W_f \tilde{x}_{l,f} + b_f \in \mathbb{R}^d,                                                                       \\
		Z       & = \left[\, z_{l,f} \,\right]_{l \in \mathcal{L},\, 1 \le f \le F} \in \mathbb{R}^{|\mathcal{L}| \times d \times F},
	\end{aligned}
\end{equation}
where $W_f \in \mathbb{R}^{d \times 1}$ and $b_f \in \mathbb{R}^{d}$ are respectively learnable projection matrices and bias parameters, shared across the lags of $\mathcal{L}$ so that a feature is encoded by the same projection at every lag.
The embedding of the whole view $f$ is denoted $z_f = \left[\, z_{l,f} \,\right]_{l \in \mathcal{L}} \in \mathbb{R}^{|\mathcal{L}| \times d}$.

An embedding dropout prevents the encoder from relying on a subset of the views.
Unlike standard dropout~\cite{dropout}, which deactivates individual units, the proposed mechanism operates at the granularity of a view, so that an entire feature is withheld from the encoder rather than a random subset of its coordinates.
During training, the encoder discards the embedding of an exogenous feature with probability $p$:
\begin{equation}
	\hat{z}_f = r_f \cdot z_f, \quad r_f \sim \text{Bernoulli}(1-p), \quad f \neq \text{target}.\label{eq:embedding-dropout}
\end{equation}
A single mask $r_f$ is drawn per batch and per view, and is shared across all the lags of $\mathcal{L}$, so that a discarded view is absent from the entire input sequence rather than from isolated timesteps.
At inference time no view is dropped, and the embeddings are used without rescaling.
The rate is set to $p = 0.2$, a common value in practice, since a rate of $0.5$ degraded forecasting accuracy on the validation set.

\begin{figure}[t]
	\centering
	\includegraphics{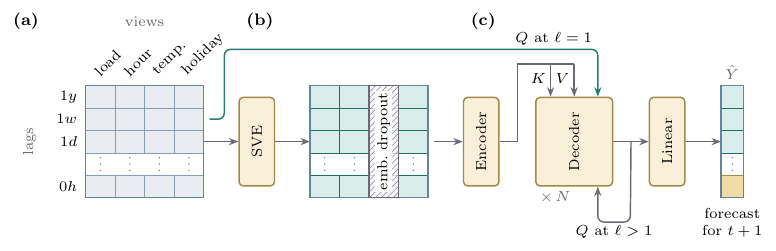}
	\caption{Single-value embedding.
		(a) The scaled time-range input gathers the rows of $\mathcal{X}$ at the lags of $\mathcal{L}$.
		(b) The SVE module applies Equation~\ref{eq:sve} to every row, so that $d_{\mathrm{model}} = F$; the hatched column is a view withheld by the embedding dropout of Equation~\ref{eq:embedding-dropout}, that is $r_f = 0$.
		(c) The decoder derives both its keys and its values from the encoded states, whereas its queries follow Equations~\ref{eq:qkv} and~\ref{eq:values}, the loop below the decoder returning $\tilde{Y}^{\ell-1}$ as the query of the next layer; the output entry attached to the null lag predicts $t+1$.
	}
	\label{fig:sve}
\end{figure}

Embeddings can be aggregated via 3 methods
\begin{itemize}
	\item Additive: $\tilde{Z} = \sum_{f=1}^F \hat{z}_f \in \mathbb{R}^{|\mathcal{L}| \times d}$,
	\item Concatenative: $\tilde{Z} = \text{Concat}(\hat{z}_1, \dots, \hat{z}_F) \in \mathbb{R}^{|\mathcal{L}| \times (d \cdot F)}$,
	\item Single-value embedding (SVE): a special case of the concatenative method, where $d=1$, hence $\tilde{Z} \in \mathbb{R}^{|\mathcal{L}| \times F}$.
\end{itemize}
The three methods therefore differ by the width of the aggregated representation, $\tilde{Z} \in \mathbb{R}^{|\mathcal{L}| \times d_{\mathrm{model}}}$, with
\begin{equation}\label{eq:dmodel}
	d_{\mathrm{model}} =
	\begin{cases}
		d,         & \text{additive}      \\
		d \cdot F, & \text{concatenative} \\
		F,         & \text{SVE}.
	\end{cases}
\end{equation}
The SVE variant admits a closed form: each $W_f$ reduces to a scalar $w_f$, so that the aggregation rescales each view by a single weight,
\begin{equation}\label{eq:sve}
	\tilde{Z}_{l,\cdot} = \tilde{x}_{l,\cdot} \odot w + b, \quad l \in \mathcal{L},
\end{equation}
where $\odot$ is the elementwise product, $w = \left(w_1, \dots, w_F\right)^\top$ gathers the weights of the $F$ views and $b = \left(b_1, \dots, b_F\right)^\top$ their biases.
The equation is written at inference, where no view is dropped.
A view is therefore summarized by one weight, shared by all its lags, and the lag axis is left untouched.
Figure~\ref{fig:sve} summarizes the whole pipeline, from the scaled time-range input to the forecast.

The aggregated embeddings $\tilde{Z}$ are first encoded by a transformer encoder $E(\cdot)$, yielding contextualized states
\begin{equation}\label{eq:encoder}
	H = E(\tilde{Z}) \in \mathbb{R}^{|\mathcal{L}| \times d_{\mathrm{model}}}.
\end{equation}
No positional encoding is applied along the lag axis: each entry is the embedding of a value observed at a fixed lag, so the anchoring time of every position is carried by the embedded values themselves, and the ordering of $\mathcal{L}$ needs no auxiliary encoding for the encoder to tell the anchors apart.
Each decoder layer $D^{\ell}(\cdot)$ performs cross-attention over $H$ to output the decoded information $\tilde{Y}^{\ell} \in \mathbb{R}^{|\mathcal{L}| \times d_{\mathrm{model}}}$, using queries, keys, and values defined as:
\begin{equation}\label{eq:qkv}
	Q^{\ell} = U^{\ell} W_Q^{\ell}, \quad K^{\ell} = H W_K^{\ell}, \quad V^{\ell} = H W_V^{\ell},
\end{equation}
where $W_Q^{\ell}, W_K^{\ell}, W_V^{\ell}$ are learnable projection matrices and
\begin{equation}\label{eq:values}
	U^{\ell} =
	\begin{cases}
		\tilde{X}_t,        & \ell = 1  \\
		\tilde{Y}^{\ell-1}, & \ell > 1.
	\end{cases}
\end{equation}
This follows the usual cross-attention layout: the decoder attends, at every layer, to the fixed context $H$ supplied by the encoder, deriving its keys and values from it, while the queries are updated layer after layer from the decoder's own evolving state $U^{\ell}$.
At the first layer the queries are the raw lagged observations $\tilde{X}_t$, so the attention weights are computed directly from the observed measurements instead of from an embedded representation.
Since $\tilde{X}_t$ has $F$ columns, $W_Q^{1}$ is the only projection mapping from $F$ rather than from $d_{\mathrm{model}}$.

The output of the last decoder layer is finally projected to the forecast:
\begin{equation}\label{eq:output}
	\hat{Y} = \text{Linear}\left(\tilde{Y}^{N}\right) \in \mathbb{R}^{|\mathcal{L}|},
\end{equation}
where $N$ is the number of decoder layers and $d_{\mathrm{model}}$ is given by Equation~\ref{eq:dmodel}.
The output layer maps each row of $\tilde{Y}^{N}$ to a single value, so that the entry $\hat{Y}_l$ predicts the observation that immediately follows the anchor $t-l$.
The forecast issued at time $t$ is the entry attached to the null lag; the remaining entries are supervised in the same way during training, which yields one target per anchor instead of a single one per input.
A horizon $\tau > 1$ is then covered autoregressively, by appending the forecast to the target series and rebuilding the scaled time-range input, so that the anchoring condition governs which entries remain observed.

\section{Experimental Settings}\label{sec:experimental-settings}
In this section, we detail the experimental setup, including the datasets used, as well as the procedures for the models training and testing phases.

\subsection{Datasets}\label{subsec:dataset}

In our experiments, we use three distinct datasets, each containing hourly time series data for the following features: load (in MW), temperature (in °C), dew point temperature (in °C), wind speed (in km/h), relative humidity (in \%), rainfall precipitation (in mm), holiday indicators (id), and school periods (boolean).

The Independent Electricity System Operator (IESO), based in Ontario, Canada, oversees the province's power system and provides key services, including real-time system operations and long-term energy planning.
The IESO dataset consists of hourly load demand data for Ontario, Canada, covering the period from 1 May 2002 to 30 September 2024, and is publicly available.\footnote{https://www.ieso.ca/power-data}

Calendar covariates have been part of hourly load models since their early formulations, where separate treatment of the day of the week and of holidays accounts for the deviations that a purely seasonal term cannot represent~\cite{semiparametric-load,hourly-load-covariates}.
Calendar features for these datasets were derived from the Python package~\emph{holidays}: a binary indicator is set to one when the date is a public holiday of the corresponding region, and a categorical identifier is taken from the mapping that the package returns for that region.
Past and future covariates can also be generated automatically from the time index, either through predefined encoders or through user-defined functions~\cite{darts}, which makes it possible to derive several calendar and positional features consistently across the training, validation and testing windows.
Weather features for IESO were retrieved from the Canadian historical climate service with one request per region and per year.\footnote{https://www.climate.weather.gc.ca/}
The Toronto station is used for the IESO dataset.

The Centro Nacional de Despacho (CND) coordinates operations and transactions within Panama’s Wholesale Electricity Market.
The CND dataset covers the national hourly load demand from 3 January 2015 to 27 June 2020 and provides the calendar and weather covariates used here, with the exception of dew point temperature~\cite{panama}.

The American Electric Power (AEP) dataset contains hourly aggregate load demand in the AEP zone of the Pennsylvania--New Jersey--Maryland (PJM) Interconnection, covering parts of the eastern and midwestern United States.
The AEP load series was obtained from the PJM energy-consumption collection through its Kaggle distribution~\cite{pjm}.
The processed AEP series covers the period from 1 October 2004 to 3 August 2018.
Weather covariates for AEP are obtained from the Open-Meteo API\footnote{https://open-meteo.com/}, using Columbus, Ohio as a regional weather proxy for the AEP service territory.

Table~\ref{tab:datasets} summarises, for each dataset, the origin of the load series and of the covariates derived from it.

\begin{table}[htb]
	\centering
	\small
	\caption{Origin and temporal coverage of the series and covariates. Calendar features are derived from the Python \emph{holidays} package, whose region code is reported here, and yield a binary holiday indicator together with a categorical holiday identifier. IESO weather features are retrieved from climate.weather.gc.ca, whereas AEP weather features are retrieved from Open-Meteo, using the location reported here.}
	\label{tab:datasets}
	\begin{tabularx}{\linewidth}{lX X X l l}
		\toprule
		Dataset & Load series                                          & Calendar & Weather source                                 & Start      & End        \\
		\midrule
		IESO    & \url{ieso.ca/Power-Data}                             & CA-ON    & Environment and Climate Change Canada, Toronto & 2002-05-01 & 2024-09-30 \\
		CND     & \url{doi.org/10.17632/byx7sztj59.1}                  & provided & provided                                       & 2015-01-03 & 2020-06-27 \\
		AEP     & \url{https://www.kaggle.com/datasets/robikscube/hourly-energy-consumption} & US-OH    & Open-Meteo, Columbus                           & 2004-10-01 & 2018-08-03 \\
		\bottomrule
	\end{tabularx}
\end{table}

\subsection{Training and Testing}\label{subsec:training-testing}
Each dataset is split chronologically into three contiguous blocks: the most recent complete year is held out for testing, or the partial final year when the series ends mid-year (as for CND, which ends in June 2020); the preceding year is used for validation; and the four years preceding the validation year are used for training, or all the remaining earlier data when fewer are available.
To preserve the temporal structure of the dataset, no shuffling is applied during the training, validation and testing phase.
Under this split, no observation posterior to a test window is available to the model at any stage.
This choice matters for the interpretation of the scores reported below, since figures obtained under different splitting strategies are not necessarily comparable.
All models in this study are evaluated under this same protocol.
All models, the proposed one and the baselines alike, are trained with the MAPE of Equation~\ref{eq:mape} as the loss function, using the AdamW optimizer, an initial learning rate of $0.001$ and a CosineAnnealing scheduler, over 300 epochs and with a batch size of 32.
Their forecasts are then evaluated on horizons of one, two and seven days under the MAPE of Equation~\ref{eq:mape}.
The comparison covers TiDE~\cite{tide}, N-BEATS~\cite{nbeats}, DLinear~\cite{dlinear} and PatchTST~\cite{patchtst}. TiDE, N-BEATS and DLinear are taken from the \emph{Darts}~\cite{darts} implementation, PatchTST from the Hugging Face backend.
All neural models use an output length of 24 hourly values. Consequently, the one-day horizon is produced directly by each model, whereas the two-day and seven-day horizons combine the direct 24-hour output with autoregressive continuation. This common output configuration keeps the comparison consistent while allowing the models to produce the first day without autoregressive rollout.

Four years of hourly data give $4 \times 8760 \approx 35{,}000$ timesteps per series, each holding several values across the views, so that the training set comprises a few hundred thousand values.

The configurations were selected through systematic hyperparameter tuning, cross-validation, and experimentation with various settings. Where applicable, the attention-based configurations use the same number of attention heads and the same representation dimension as the proposed model. The remaining architectural settings follow the design of each model rather than enforcing an exactly matched parameter count.

The proposed model is implemented in Python using PyTorch 2.12 with CUDA 13.0, and the experiments are performed on the Université du Québec à Montréal (UQAM) university cluster using an NVIDIA A30 GPU and an Intel Xeon Gold 6338 CPU at 2.00~GHz.

\subsection{Hyperparameters}\label{subsec:hyperparameters}
Table~\ref{tab:hyperparameters} reports the configuration of the deep baselines and of the proposed model, including their shared optimizer, scheduler, and training settings.
The baselines use a contiguous look-back of $168$ hours, corresponding to a full weekly cycle. This follows the standard input design of these architectures and provides a common temporal context for the comparison.

\begin{table}[htb]
	\centering
	\small
	\caption{Hyperparameters of the deep models. The upper block gathers the settings shared by every model, the lower one those specific to an architecture. A dash indicates a hyperparameter that does not apply to the architecture. The lag set $\mathcal{L}$ is defined in Section~\ref{sec:proposed-approach}.}
	\label{tab:hyperparameters}
	\begin{tabular}{lccccc}
		\toprule
		                       & SVE           & N-BEATS                                         & TiDE & DLinear & PatchTST \\
		\midrule
		Epochs                 & \multicolumn{5}{c}{300}                                                                     \\
		Loss                   & \multicolumn{5}{c}{MAPE}                                                                     \\
		Timestep               & \multicolumn{5}{c}{1 hour}                                                                  \\
		Covariates             & \multicolumn{5}{c}{time and exogenous features}                                             \\
		Optimiser              & \multicolumn{5}{c}{AdamW}                                                                   \\
		Scheduler              & \multicolumn{5}{c}{CosineAnnealingLR}                                                       \\
		Learning rate          & \multicolumn{5}{c}{0.001}                                                                   \\
		Weight decay           & \multicolumn{5}{c}{0.00001}                                                                 \\
		Dropout                & \multicolumn{5}{c}{0.1}                                                                     \\
		\midrule
		Input length           & $\mathcal{L}$ & 168  & 168     & 168     & 168      \\
		Output length          & 24                                                    & 24   & 24            & 24      & 24       \\
		Stacks/blocks/widths   & ---           & 3/3/[16, 16, 16]                         & ---   & ---     & ---      \\
		Encoder/decoder layers & 2/2           & ---                                    & 2/2   & ---     & 4/-      \\
		Hidden size            & 32            & ---            & 32            & ---     & 32       \\
		Attention heads        & 2             & ---                                    & ---   & ---     & 2        \\
		Num params             & 33,605        & 200,635       & 287,046      & 346,818  & 104,408    \\
		\bottomrule
	\end{tabular}
\end{table}

\section{Experimental Results}\label{sec:experimental-results}
\begin{figure}[h]
	\begin{minipage}[b]{0.33\linewidth}
		\centering
		\centerline{\includegraphics[width=\linewidth]{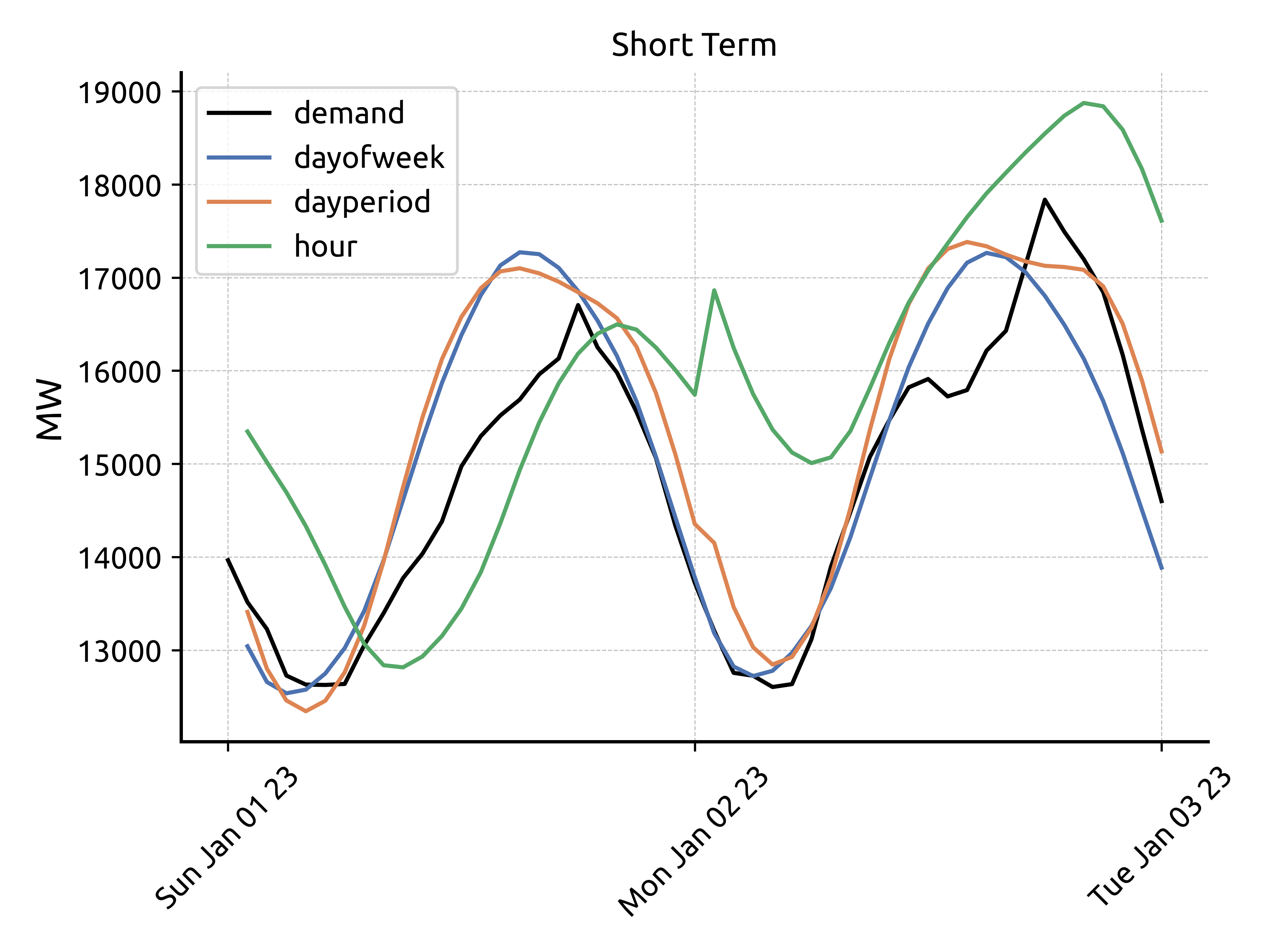}}
		\centerline{\includegraphics[width=\linewidth]{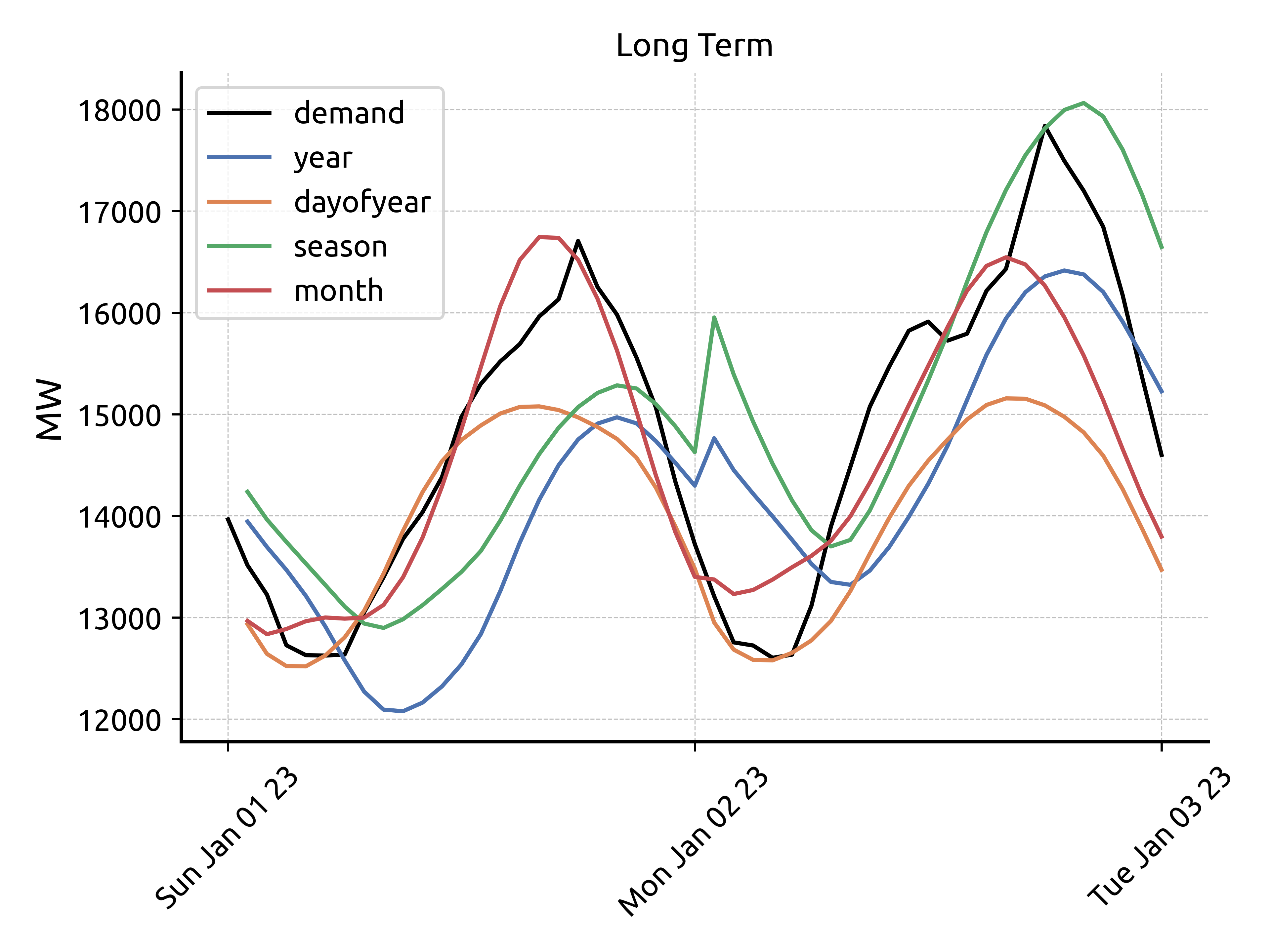}}
	\end{minipage}%
	\begin{minipage}[b]{0.33\linewidth}
		\centering
		\centerline{\includegraphics[width=\linewidth]{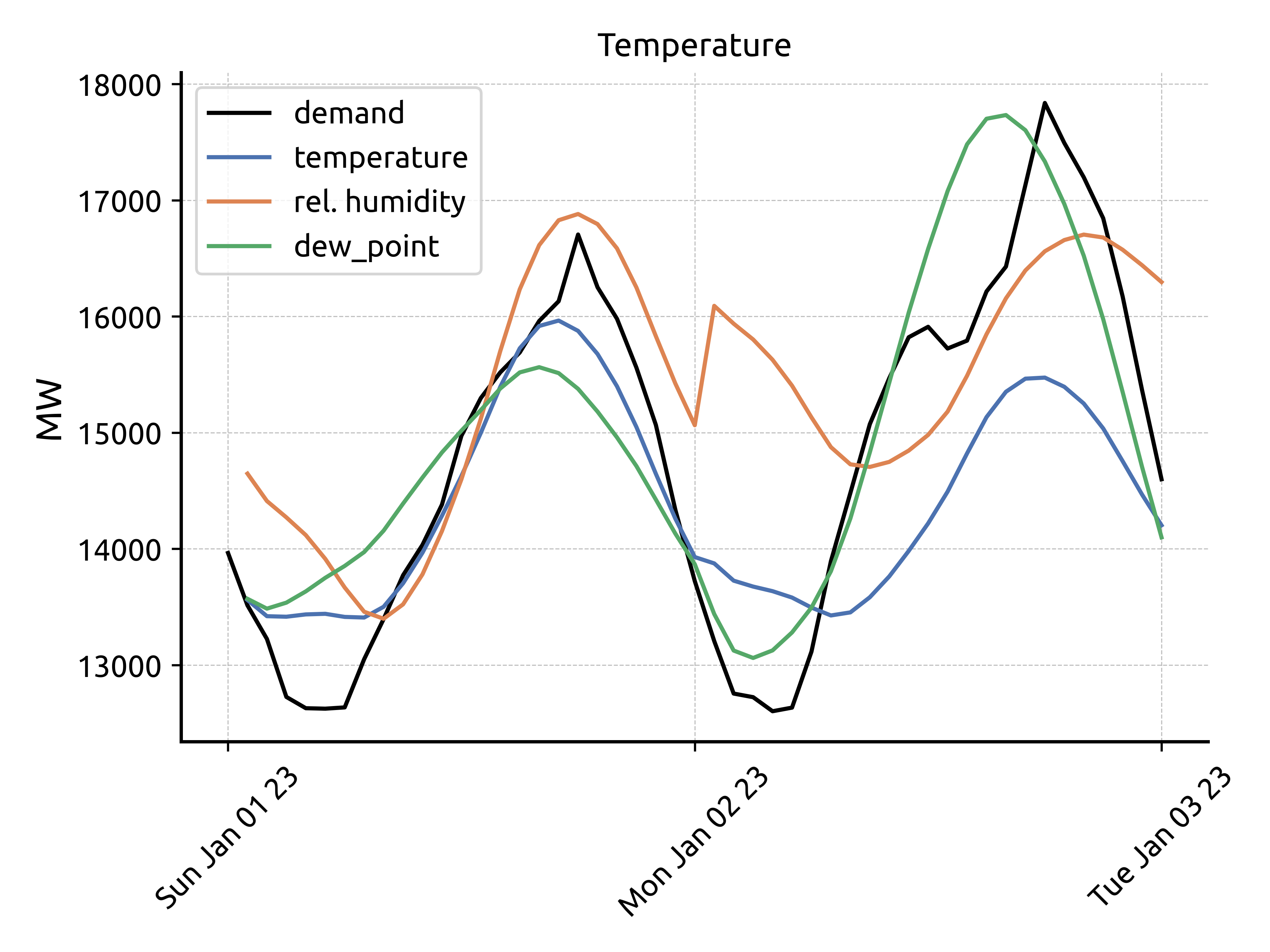}}
		\centerline{\includegraphics[width=\linewidth]{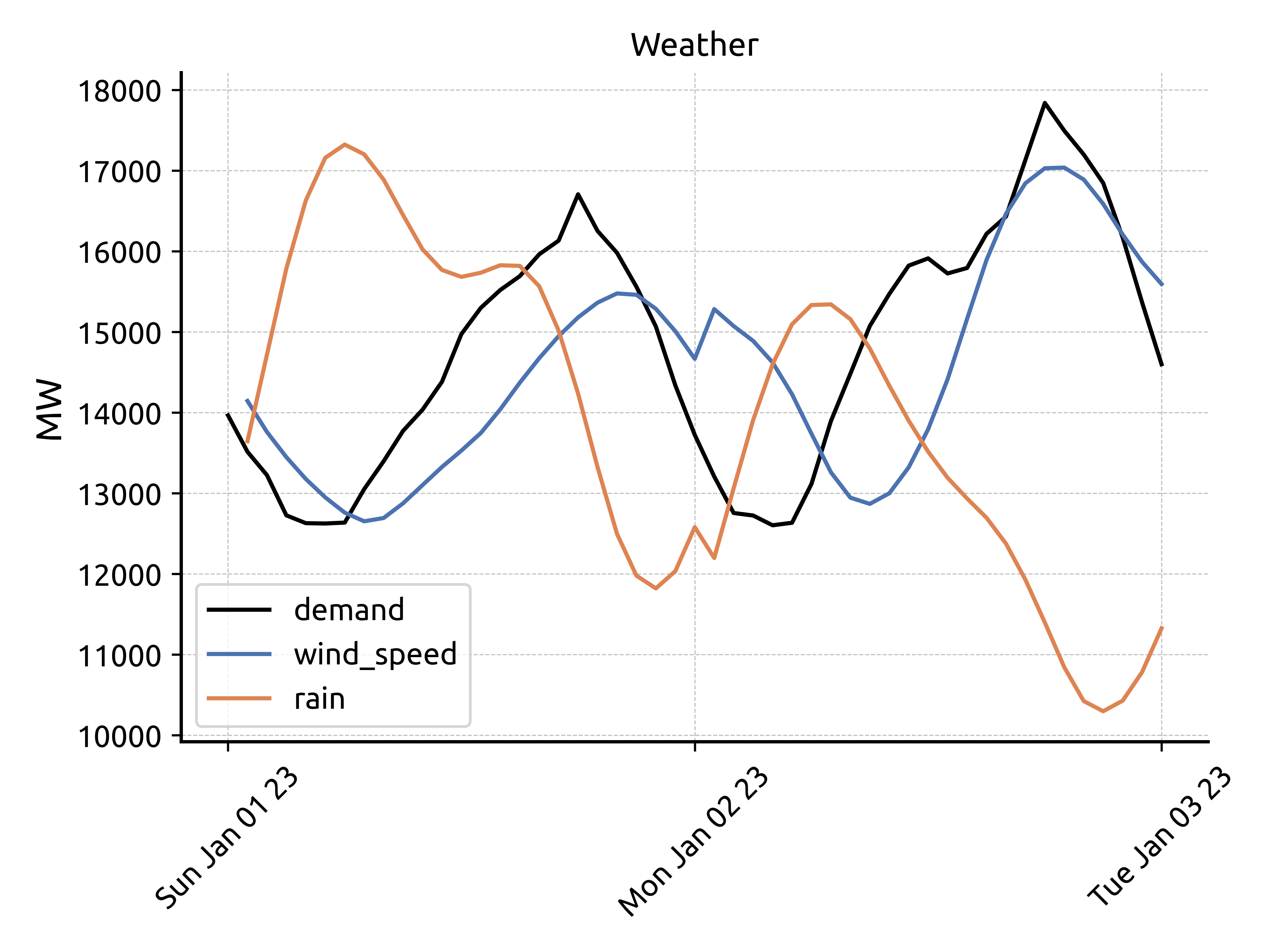}}
	\end{minipage}
	\begin{minipage}[b]{0.33\linewidth}
		\centering
		\centerline{\includegraphics[width=\linewidth]{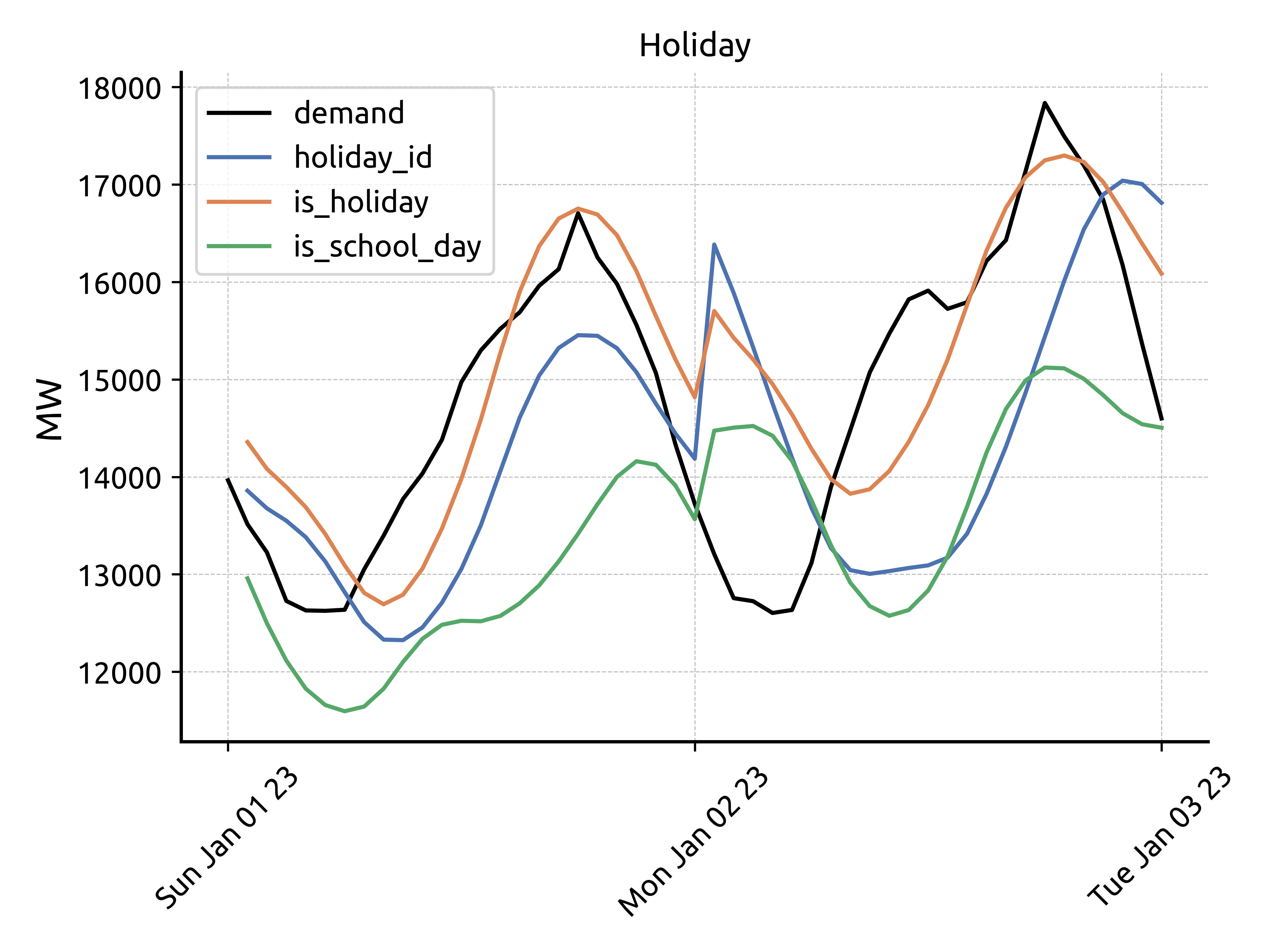}}
		\centerline{\includegraphics[width=\linewidth]{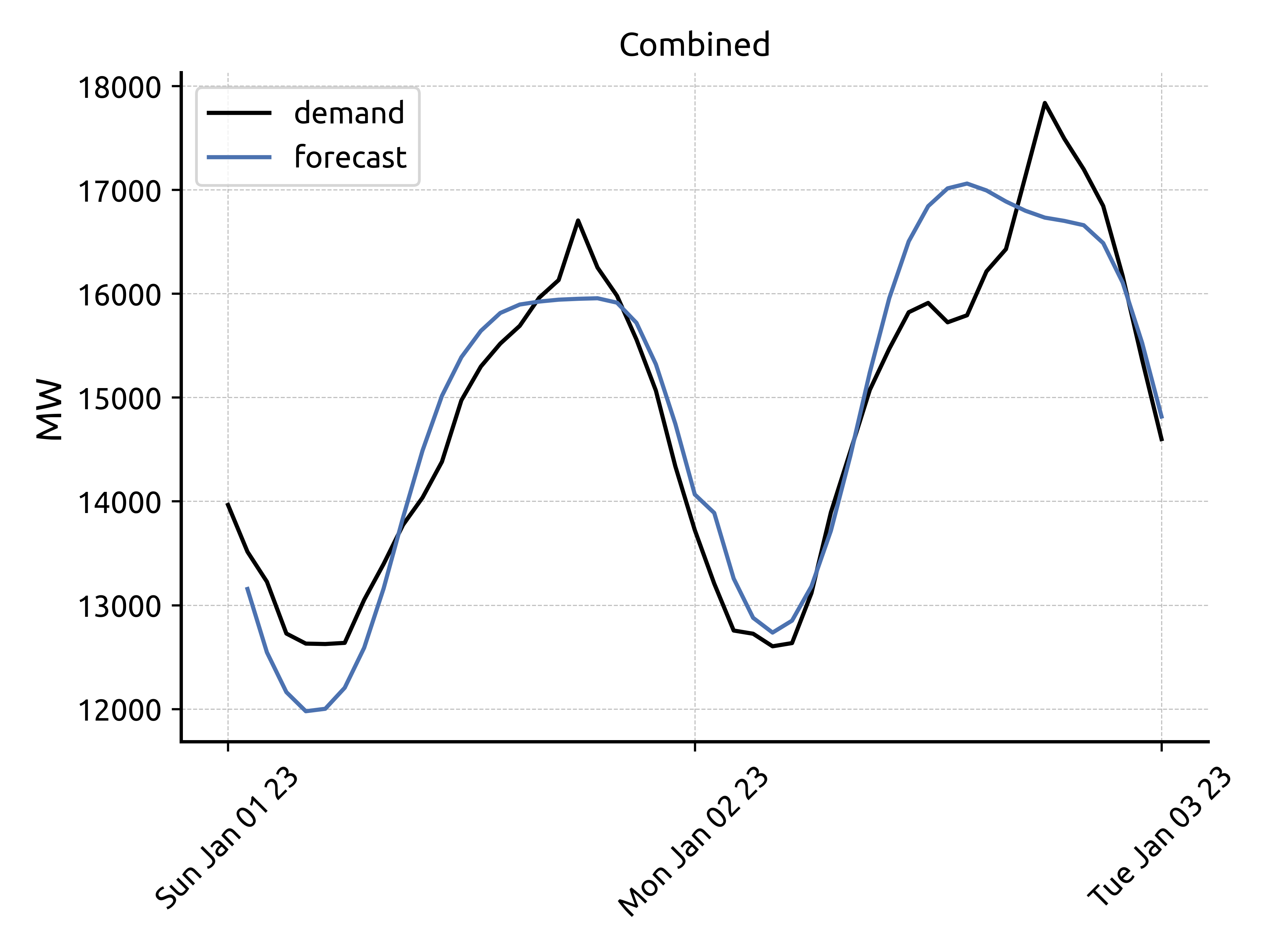}}
	\end{minipage}
	\caption{Embeddings effects of long term, short term, temperature, weather and holidays related views on the IESO dataset for a 48 hours horizon using the SVE method.
		Each panel reports the forecast obtained when a single view is retained at inference, in colour, against the observed load, in black.
		The combined effect is shown at the bottom right.
		The horizontal axis is the forecast horizon in hours and the vertical axis is the load in MW.
	}
	\label{fig:plot-demand}
\end{figure}

The embedding dropout is a way to prevent the model from relying too much on some embeddings.
An intriguing aspect of the embedding dropout technique is its ability to isolate the effect of a single view by selectively omitting all others.
This ablation is deterministic and applied at inference only: the mask $r_f$ of Equation~\ref{eq:embedding-dropout} is fixed so that a single view is retained, the model itself remaining unchanged.
Figure~\ref{fig:plot-demand} illustrates the embedding effects on the IESO load dataset.

\begin{figure}[h]
	\centering
	\includegraphics[width=0.75\linewidth]{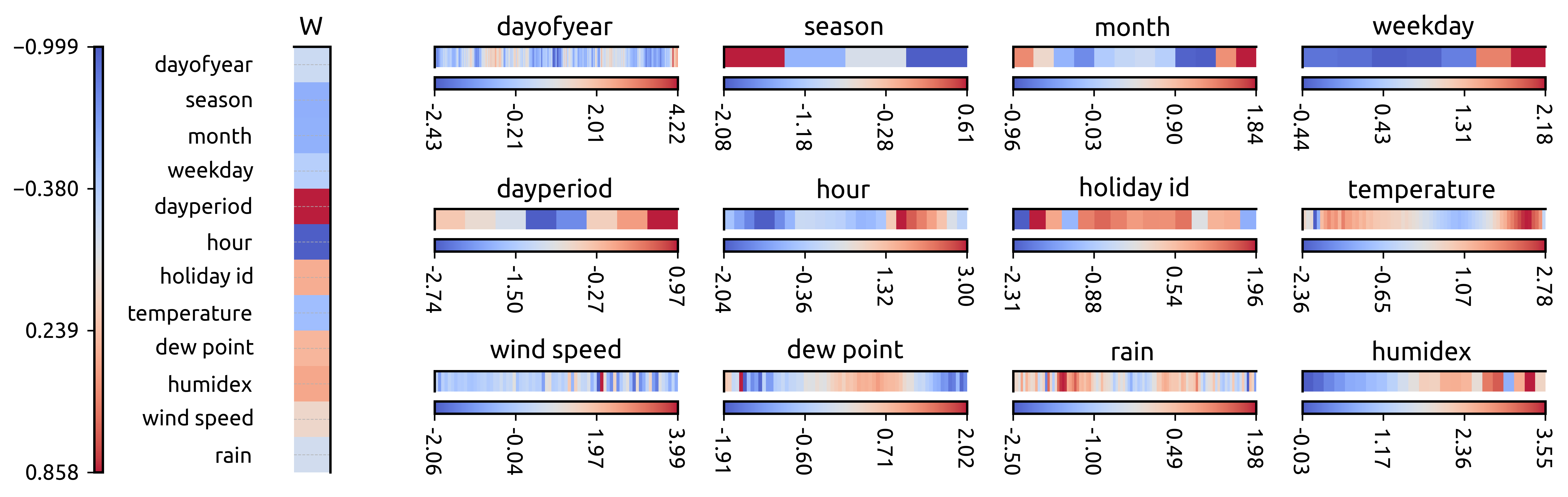}
	\caption{Embedded encoder representations. Each row corresponds to a view and each cell to one of its discrete slots, continuous values being rounded to the nearest integer: the hour view, for instance, is discretised into twenty-four slots, the first of which carries the weight assigned to midnight. Leftmost cells represent the lowest categorical values (e.g., Spring to Winter for season). Cell intensity encodes the learned embedding weight on a diverging scale centred on zero. The weighting vector $W$ gives, for each view, the weight it carries in the forecast.}
	\label{fig:net-embeddings}
\end{figure}

Given that the SVE method rounds input data to the nearest integer, examining the learned embedding weights provides insights into the model’s internal behavior.
Figure~\ref{fig:net-embeddings} illustrates the learned embeddings across the encoder’s input views.

Forecasting can vary in complexity depending on the context.
For instance, in the energy sector, forecasting load becomes particularly challenging during holidays due to irregular usage patterns and deviations from typical demand trends.
Additionally, in certain scenarios, datasets may exhibit sparsity, often caused by missing data due to malfunctioning sensors or data acquisition failures.

The behavior of the proposed approach is examined on both holiday periods and corrupted datasets.
Corruption emulates a faulty sensor: each exogenous feature of a row is replaced, with probability 50\%, by its minimum, maximum or mean, these being the values that an acquisition pipeline typically returns or imputes when a sensor fails.
Table~\ref{tab:mape} presents the MAPE scores for the experimented embedding methods, comparing performance on a forecasting horizon of one day (1d), two days (2d), and seven days (7d) across the entire test set, the subset containing holiday periods, and the corrupted subset.
Table~\ref{tab:vs} compares our approach with the selected neural baselines on the three datasets.

\begin{table}[htb]
	\caption{MAPE scores for different embeddings combination methods. Parameter counts include every trainable weight of the network, embedding layers comprised.}
	\begin{center}
		\scriptsize
		\setlength{\tabcolsep}{2.5pt}
		\renewcommand{\arraystretch}{1.2}
		\begin{tabular}{ccccc|ccc|ccc}
			\toprule[1.0pt]
			\multicolumn{1}{c}{}                            & \multicolumn{10}{c}{\textbf{Embeddings methods}}                                                                                                                                                    \\
			\multicolumn{2}{c}{}                            & \multicolumn{3}{c}{\textbf{additive}}            & \multicolumn{3}{c}{\textbf{concatenative}} & \multicolumn{3}{c}{\textbf{SVE}}                                                                    \\
			\multicolumn{2}{c}{num params}                  & \multicolumn{3}{c}{93k}                          & \multicolumn{3}{c}{143k}                   & \multicolumn{3}{c}{33,605}                                                                          \\
			\hline
			\multicolumn{2}{c}{horizon}                     & 1d                                               & 2d                                         & 7d                               & 1d   & 2d            & 7d   & 1d   & 2d            & 7d          \\
			\hline
			\multirow{3}{*}{\rotatebox[origin=c]{90}{IESO}} & full                                             & \textbf{4.41} & \textbf{4.96} & \textbf{5.77} & 5.35 & 5.81 & 6.54 & 4.64 & 5.17 & 6.24 \\
			                                                & holidays                                         & 5.16                                       & 6.10                             & 6.68 & \textbf{4.77} & \textbf{5.49} & \textbf{6.32} & 5.55 & 6.33 & 7.39 \\
			                                                & corrupted                                        & 6.08                                       & 6.92                             & 8.20 & \textbf{5.25} & \textbf{5.98} & \textbf{7.04} & 5.84 & 6.64 & 7.78 \\
			\hline
			\multirow{3}{*}{\rotatebox[origin=c]{90}{CND}}  & full                                             & 6.91                                       & 7.20                             & \textbf{7.55} & 6.67 & \textbf{7.20} & 7.80 & \textbf{6.65} & 7.37 & 8.21 \\
			                                                & holidays                                         & 6.63                                       & 7.78                             & 10.22 & \textbf{6.62} & \textbf{7.63} & \textbf{9.54} & 7.24 & 8.42 & 10.33 \\
			                                                & corrupted                                        & 6.91                                       & 8.06                             & 10.67 & \textbf{6.78} & \textbf{7.75} & \textbf{9.80} & 7.37 & 8.63 & 10.71 \\
			\hline
			\multirow{3}{*}{\rotatebox[origin=c]{90}{AEP}}  & full                                             & \textbf{4.59} & \textbf{5.08} & \textbf{5.74} & 8.64 & 8.94 & 9.31 & 9.69 & 9.93 & 10.24 \\
			                                                & holidays                                         & \textbf{8.69} & \textbf{8.37} & \textbf{8.80} & 9.67 & 9.49 & 9.48 & 9.31 & 9.56 & 9.85 \\
			                                                & corrupted                                        & \textbf{10.32} & \textbf{9.89} & \textbf{10.82} & 11.14 & 11.20 & 11.11 & 10.93 & 11.45 & 11.81 \\
			\hline
		\end{tabular}
		\label{tab:mape}
	\end{center}
\end{table}

\begin{table}[htb]
	\centering
	\caption{Multivariate time series forecasting results on three public datasets. Parameter counts follow the convention of Table~\ref{tab:mape}.}
	\scriptsize
	\renewcommand{\arraystretch}{1.2}
	\begin{tabularx}{\linewidth}{cc*{5}{>{\centering\arraybackslash}X}}
		\toprule[1.0pt]
		\multicolumn{2}{c}{Models}     & \multicolumn{1}{c}{SVE}      &
		\multicolumn{1}{c}{N-BEATS}  & \multicolumn{1}{c}{TiDE}       &
		\multicolumn{1}{c}{DLinear}    & \multicolumn{1}{c}{PatchTST}                                                                          \\
		\multicolumn{2}{c}{num params} & \multicolumn{1}{c}{33,605}   &
		\multicolumn{1}{c}{200,635}  & \multicolumn{1}{c}{287,046}    & \multicolumn{1}{c}{346,818} & \multicolumn{1}{c}{104,408}                  \\
		\hline
		\multirow{3}{*}{\rotatebox[origin=c]{90}{IESO}}
		 & 1d                           & 4.64                    & 4.62 & 3.49 & \textbf{2.92} & 4.11 \\
		 & 2d                           & 5.17                    & 5.50 & 4.36 & \textbf{3.62} & 4.83 \\
		 & 7d                           & 6.24                    & 7.33 & 5.61 & \textbf{4.92} & 5.71 \\
		\hline
		\multirow{3}{*}{\rotatebox[origin=c]{90}{CND}}
		 & 1d                           & 6.65                    & 5.04 & 3.94 & \textbf{3.69} & 4.53 \\
		 & 2d                           & 7.37                    & 5.48 & 4.58 & \textbf{4.32} & 4.89 \\
		 & 7d                           & 8.21                    & 6.08 & 5.34 & 5.35 & \textbf{5.32} \\
		\hline
		\multirow{3}{*}{\rotatebox[origin=c]{90}{AEP}}
		 & 1d                           & 9.69                    & 8.47 & 4.29 & \textbf{3.49} & 5.89 \\
		 & 2d                           & 9.93                    & 9.37 & 5.73 & \textbf{4.64} & 7.17 \\
		 & 7d                           & 10.24                  & 11.19 & 8.25 & \textbf{7.14} & 8.77 \\
		\hline
	\end{tabularx}
	\label{tab:vs}
\end{table}

\begin{figure}[htb]
	\begin{minipage}[b]{0.33\linewidth}
		\centering
		\centerline{\includegraphics[width=\linewidth]{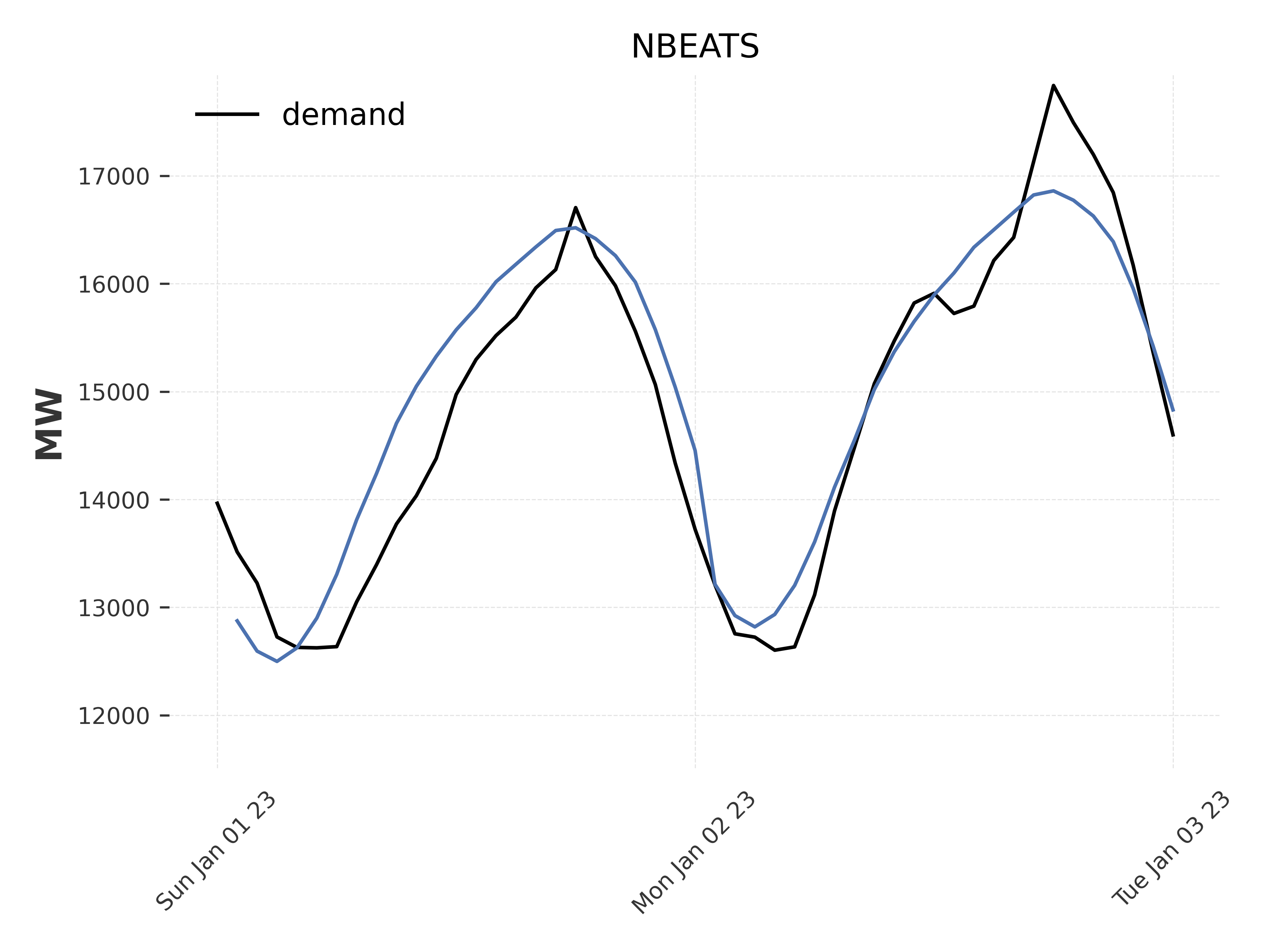}}
		\centerline{\includegraphics[width=\linewidth]{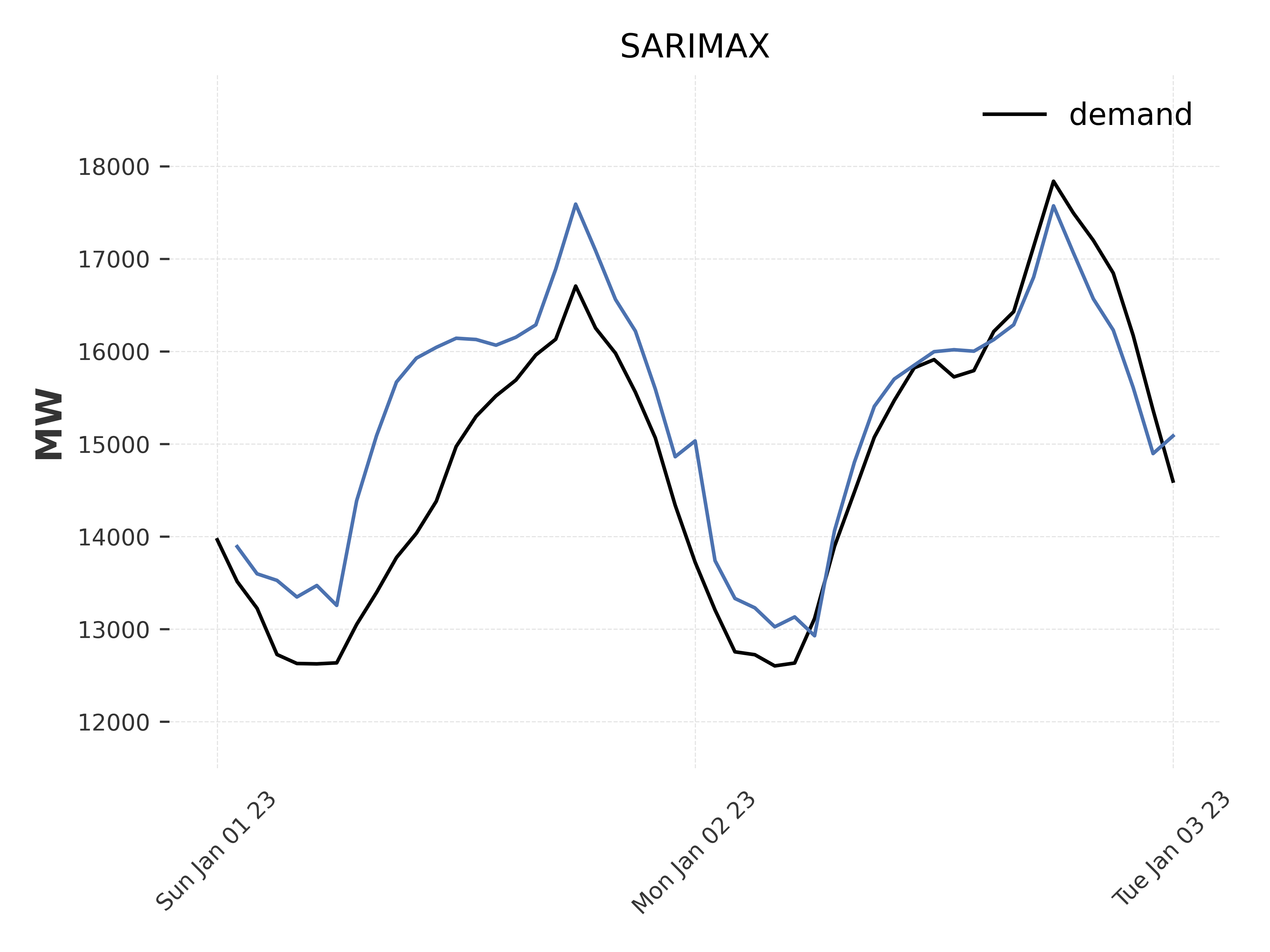}}
	\end{minipage}%
	\begin{minipage}[b]{0.33\linewidth}
		\centering
		\centerline{\includegraphics[width=\linewidth]{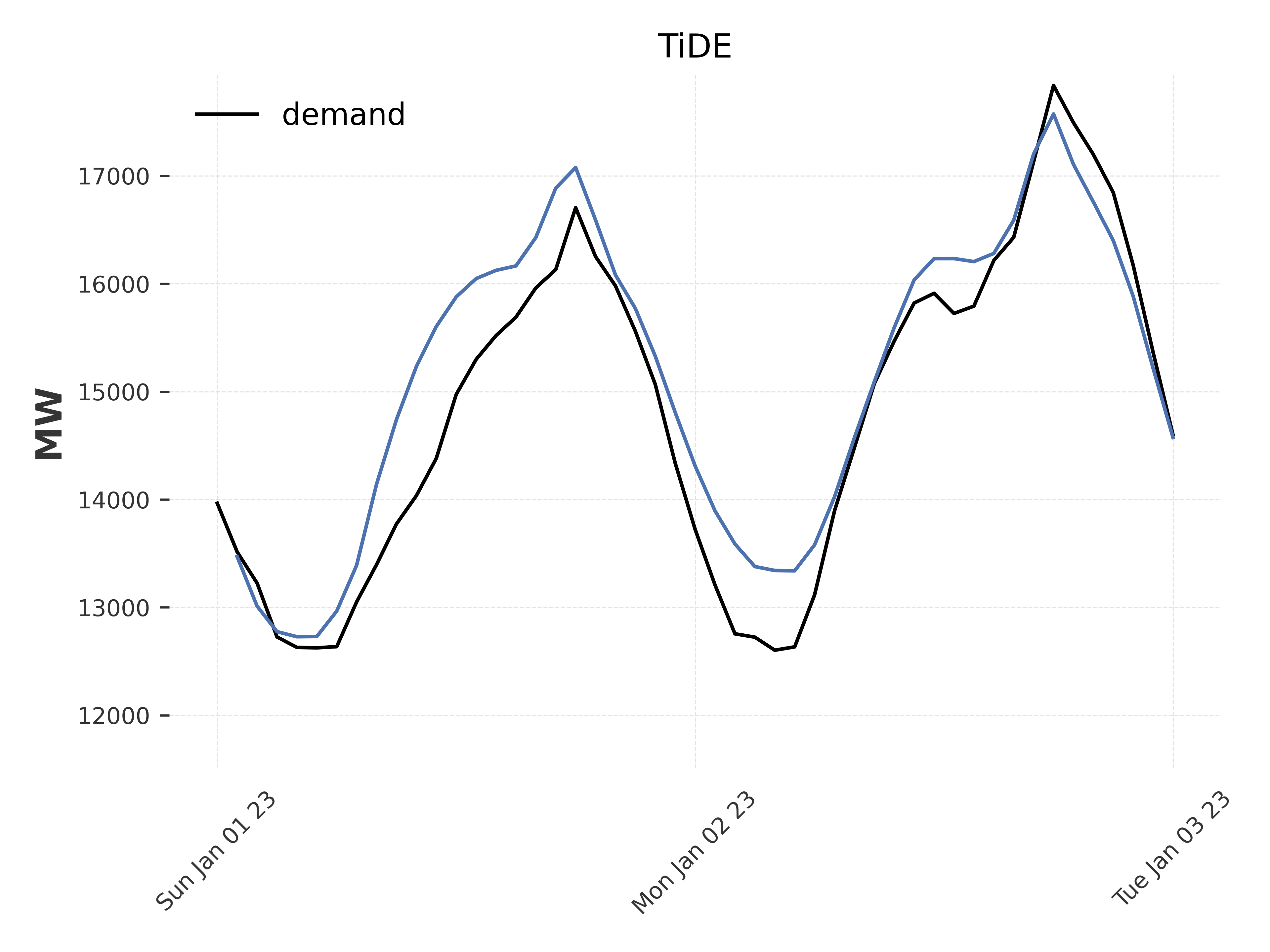}}
		\centerline{\includegraphics[width=\linewidth]{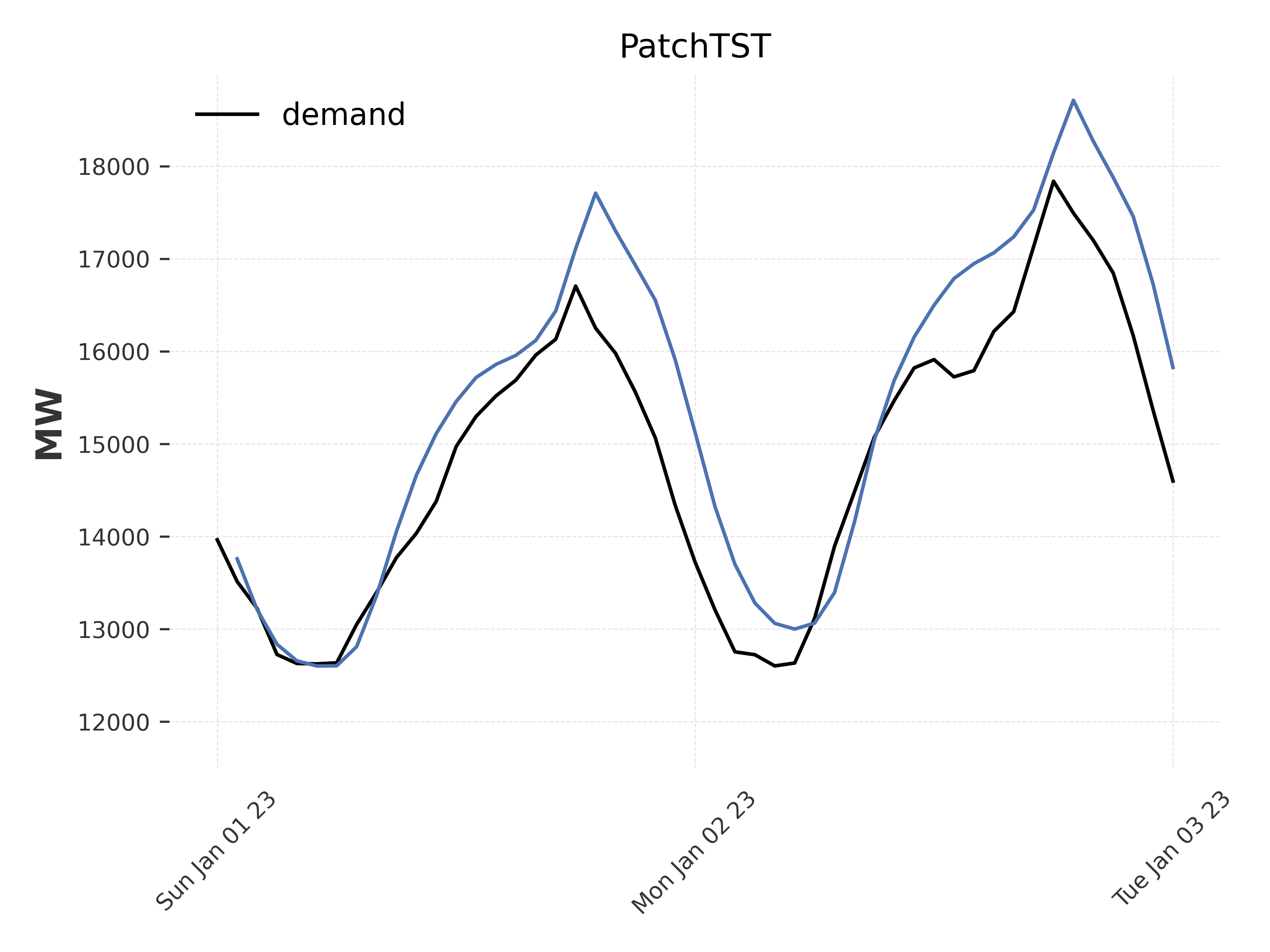}}
	\end{minipage}%
	\begin{minipage}[b]{0.33\linewidth}
		\centering
		\centerline{\includegraphics[width=\linewidth]{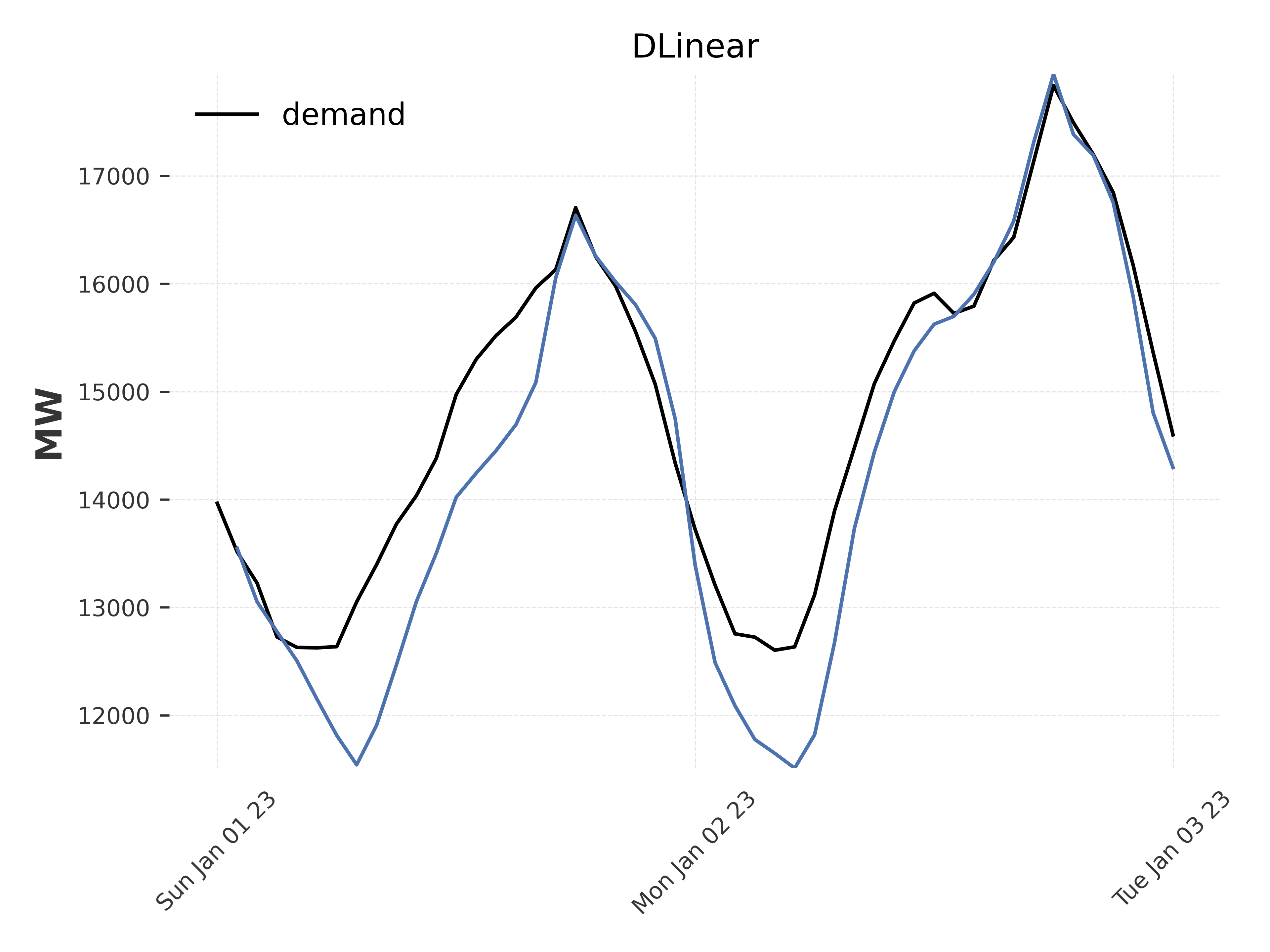}}
		\centerline{\includegraphics[width=\linewidth]{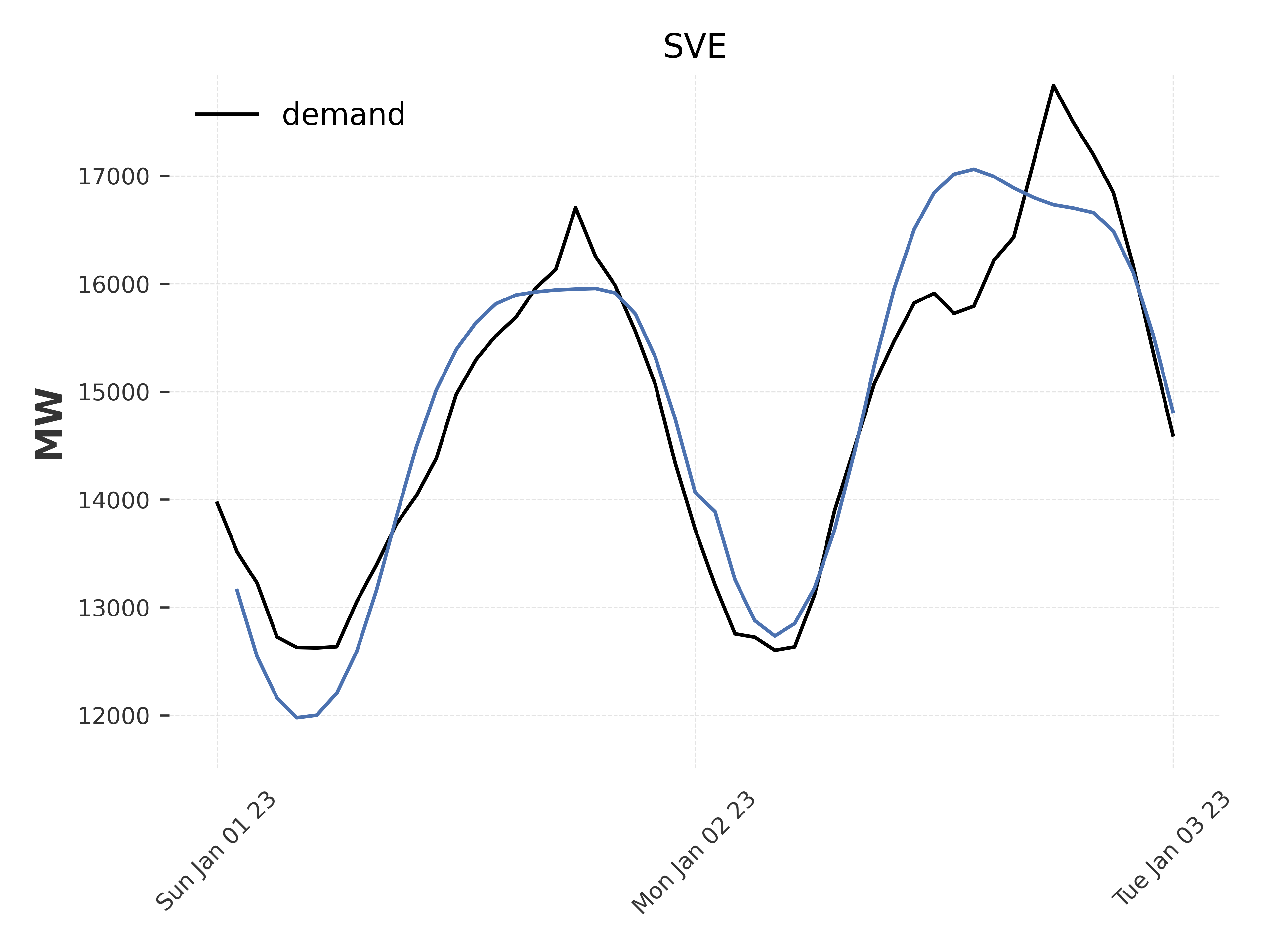}}
	\end{minipage}
	\caption{Forecasts from N-BEATS, TiDE, DLinear, SARIMAX, PatchTST and the proposed model on the IESO test set, over a two-day horizon spanning January 1st and 2nd 2023.
		Each panel reports the forecast of one model, in colour, against the observed load, in black.
		The proposed model is shown at the bottom right.
		The horizontal axis is the forecast horizon in hours and the vertical axis is the load in MW.
		All models are given the same forecast origin and the same exogenous features, and the period covers a public holiday, when the daily profile departs from the usual weekday pattern.
		SARIMAX is shown for reference only and is excluded from the comparison of Table~\ref{tab:vs}.
	}
	\label{fig:forecasts}
\end{figure}

Figure~\ref{fig:forecasts} compares the forecasts of the scored models on the IESO test set over the same two days, which span the New Year's Day 2023 public holiday, when the daily load profile departs from the usual weekday pattern.
SARIMAX, with order $(6,1,6)$ and seasonal order $(1,0,1,24)$, is included for qualitative reference only. Because it is estimated on the complete available history, its forecast is not directly comparable to the neural-model scores and is excluded from the quantitative comparison.

\section{Discussion}\label{sec:discussion}
We interpret the results of Table~\ref{tab:vs} in light of a structural property of the proposed input.
The lag set $\mathcal{L}$ embeds the weekly and yearly lags, which point to values observed before the forecast origin and remain anchored for the entire duration of a one-week horizon.
Although the short lags can be replaced by generated values during continuation beyond the direct 24-hour output, the model retains its weekly and yearly references as observed values for horizons up to one week. This property follows from the construction of $\mathcal{L}$, although the present comparison does not isolate its contribution from the rest of the architecture.

Empirically, the results vary across datasets and models. SVE increases from 4.64 to 6.24 MAPE on IESO, from 6.65 to 8.21 on CND, and from 9.69 to 10.24 on AEP between one-day and seven-day horizons. The baselines show different trends and magnitudes of change, with DLinear performing best on IESO and AEP and DLinear or PatchTST performing best on CND. Because each configuration was trained once, these differences should be interpreted descriptively rather than as evidence of statistically significant superiority. The two-day and seven-day results also include autoregressive continuation after the direct 24-hour output, so they reflect both model performance and rollout behavior.

By contrast, the four baselines cannot offer this property, because their input construction is designed for a contiguous window.
N-BEATS separates trend from seasonality in dedicated stacks, TiDE encodes the look-back window with a dense projection, DLinear decomposes it into trend and seasonal components, and PatchTST attends over patches of it; each of these operations is designed for a contiguous segment of the series with a fixed interval.
Such a window would need to span at least one year to present the one-hour, one-week, and one-year lags simultaneously, and would therefore use a substantially different input representation.
Supplying the same lag set to the baselines would not isolate the contribution of the input.
The comparison of Table~\ref{tab:vs} therefore contrasts complete designs, in which the input construction and the architecture are chosen jointly, rather than isolating the lag set alone.

However, the anchoring property also carries a cost, which the CND dataset exposes.
Its test set covers 2020, when the pandemic caused a sustained drop in consumption, while the models are trained on pre-pandemic years and no covariate carries that information.
The degradation is thus a distribution shift rather than a forecasting failure.
The shift penalises the proposed model more heavily than the baselines, because the weekly and yearly lags point to a regime that no longer holds.
The same limit applies when forecasting beyond a one-week horizon, where only the yearly lag remains anchored, a case we leave outside this study.

Two caveats qualify these results.
The proposed model is autoregressive, so early errors propagate through later steps, and the exogenous variables are presumed error-free, which is unrealistic since weather forecasts come from other models.
The corrupted-input results do not establish robustness to faulty covariates: replacing exogenous values can produce substantial degradation, especially at longer horizons and on CND. They should therefore be interpreted as a sensitivity analysis.

Beyond regularisation, the dropout doubles as an ablation mechanism, since withholding all views but one at inference exposes the contribution of each covariate.
In Figure~\ref{fig:plot-demand}, short-term features capture intraday cycles, long-term features the seasonal trend, weather features the response to environmental conditions, and holiday indicators track irregular deviations from the usual pattern.
The embeddings of Figure~\ref{fig:net-embeddings} exhibit the same distinction between categorical and continuous features, with categorical features such as season and weekday carrying strong periodic weights and continuous features such as temperature and humidity varying smoothly.
Together the two figures link the learned representations to identifiable drivers of demand.
The attention weights of the first decoder layer give a similar view, since at that layer the queries are the raw lagged observations and the weights are therefore computed directly from the measured signal. They serve as qualitative diagnostics of how the forecast is assembled, without claiming a faithful attribution.

Across the comparison, the models follow their respective architectural configurations rather than an exactly matched parameter count. The results show that scaled time-range inputs provide a compact way to emphasize specific lags, such as the previous day or the previous week, but they do not establish a uniform accuracy advantage over the baselines.

The concatenative method performs slightly better than the single-value embedding on the CND dataset.
The absence of dew point temperature there suggests that the choice of exogenous variables influences the effectiveness of the single-value method.
It remains competitive nonetheless, at a quarter of the parameter count of the concatenative variant.

Not all embeddings contribute equally, as emphasized in~\cite{camul}, and their relative importance varies across series and over time.
This suggests that an adaptive mechanism adjusting the contribution of each view could provide greater robustness than a fixed weighting.
The same reasoning applies beyond load forecasting.
Financial series lack a stable periodicity, and would require a lag policy conditioned on the state of the series rather than the fixed lag set used here.
Adaptively weighting the views and learning such a lag policy define the main directions for future work.

\section{Conclusion}\label{sec:conclusion}
This paper addressed short-term load forecasting with a compact multi-view architecture.
We proposed an encoder--decoder architecture resting on three components, namely a scaled time-range input that samples eleven lags instead of a contiguous window, a single-value embedding that encodes each view with one scalar weight, and a dropout applied at the granularity of a view.
The comparison of three aggregation strategies shows that the way embeddings are combined affects accuracy, with the best variant depending on the dataset and evaluation protocol. The single-value variant achieves this representation with substantially fewer trainable parameters than the other embedding strategies.
Beyond regularisation, the view-level dropout exposes the contribution of each covariate to the forecast trajectory, since withholding all views but one isolates that contribution and links the learned weights to identifiable drivers of demand.

Under a strictly chronological protocol, the forecasting accuracy of the proposed model varies across datasets and horizons. The proposed SVE representation is not the best-performing model in every setting, but it provides a compact alternative to the larger neural baselines. Its lag set preserves weekly and yearly references as observed values during forecasts of up to one week, although the present experiments do not isolate the contribution of this property from the rest of the architecture.
The same property carries a cost, which the CND results make explicit.
When the consumption regime changes, as during the 2020 pandemic, the anchors point to a period that no longer holds and the proposed model is penalised more than the baselines.
One limitation bounds these findings, namely that beyond a one-week horizon only the yearly lag remains anchored.

The main direction that follows from this work concerns the lag set, which is fixed whereas the lags that carry information depend on the state of the series.
Learning a lag policy conditioned on that state would adapt the input to the regime holding at the forecast origin, would keep more than one anchor available beyond a one-week horizon, and would by extension reach domains whose periodicity is not stable.

\section*{Author Contributions}
J.G.-V.: conceptualization, methodology, software, validation, formal analysis, investigation, resources, data curation, writing---original draft preparation, writing---review and editing, visualization; A.B.M. and \'E.B.: writing---review and editing, supervision, project administration. All authors have read and agreed to the published version of the manuscript.

\section*{Funding}
This research received no external funding.

\section*{Data Availability}
All three datasets used in this study are publicly available. The IESO load series is distributed by the Independent Electricity System Operator at \url{https://ieso.ca/Power-Data/Data-Directory}, the CND dataset is available at \url{https://doi.org/10.17632/byx7sztj59.1}, and the AEP load series was obtained from the PJM energy-consumption collection through its Kaggle distribution at \url{https://www.kaggle.com/datasets/robikscube/hourly-energy-consumption}. The weather covariates are derived from the Environment and Climate Change Canada archive, Open-Meteo data for AEP, and the calendar covariates from the Python \emph{holidays} package. The implementation used to produce the reported results is available at \url{https://gitlab.info.uqam.ca/guite-vinet.julien/prevdem-net-paper}.

\section*{Acknowledgments}
During the preparation of this manuscript, the authors used the opencode assistant, powered by the Big Pickle model, to correct grammatical and typographical errors, to propose alternative wordings, and to cross-check factual statements. The authors have reviewed and edited the output and take full responsibility for the content of this publication.

\section*{Conflicts of Interest}
The authors declare no conflicts of interest.

\bibliographystyle{unsrtnat}
\bibliography{bibliographie}

\end{document}